\documentclass[pdflatex,sn-mathphys-num]{sn-jnl}

\usepackage{graphicx}%
\usepackage{multirow}%
\usepackage{amsmath,amssymb,amsfonts}%
\usepackage[title]{appendix}%
\usepackage{xcolor}%
\usepackage{textcomp}%
\usepackage{manyfoot}%
\usepackage{booktabs}%
\usepackage{algorithm}%
\usepackage{algorithmicx}%
\usepackage{algpseudocode}
\usepackage{listings}%
\usepackage{amsthm}%
\DeclareMathAlphabet{\mathcal}{OMS}{cmsy}{m}{n}

\theoremstyle{thmstyleone}%
\theoremstyle{thmstyletwo}%

\theoremstyle{thmstylethree}%

\begin{document}

\title[Article Title]{Person Identification from Egocentric Human-Object Interactions using 3D Hand Pose}


\author*[1,2]{\fnm{Muhammad} \sur{Hamza}}\email{hamzac070@gmail.com}

\author[2,3]{\fnm{Danish} \sur{Hamid}}\email{danish.hamid@au.edu.pk}
\equalcont{These authors contributed equally to this work.}

\author[1,2]{\fnm{Muhammad} \sur{Tahir Akram}}\email{tahirakram15@hotmail.com}
\equalcont{These authors contributed equally to this work.}

\affil*[1]{\orgdiv{Department of Creative Technologies}, \orgname{Air University}, \orgaddress{\postcode{44230}, \state{Islamabad}, \country{Pakistan}}}

\affil[2]{\orgdiv{Department of Creative Technologies}, \orgname{Air University}, \orgaddress{\postcode{44230}, \state{Islamabad}, \country{Pakistan}}}

\affil[2]{\orgdiv{Department of Creative Technologies}, \orgname{Air University}, \orgaddress{\postcode{44230}, \state{Islamabad}, \country{Pakistan}}}


\abstract{Human-Object Interaction Recognition (HOIR) and user identification play a crucial role in advancing augmented reality (AR)-based personalized assistive technologies. These systems are increasingly being deployed in high-stakes, human-centric environments such as aircraft cockpits, aerospace maintenance, and surgical procedures. This research introduces \textit{I2S (Interact2Sign)}, a multi-stage framework designed for unobtrusive user identification through human-object interaction recognition, leveraging 3D hand pose analysis in egocentric videos. I2S utilizes handcrafted features extracted from 3D hand poses and performs sequential feature augmentation: first identifying the object class, followed by HOI recognition, and ultimately, user identification. A comprehensive feature extraction and description process was carried out for 3D hand poses, organizing the extracted features into semantically meaningful categories: \textbf{Spatial}, \textbf{Frequency}, \textbf{Kinematic}, \textbf{Orientation}, and a novel descriptor introduced in this work, the \textbf{Inter-Hand Spatial Envelope (IHSE)}. Extensive ablation studies were conducted to determine the most effective combination of features. The optimal configuration achieved an impressive average F1-score of 97.52\% for user identification, evaluated on a bimanual object manipulation dataset derived from the ARCTIC and H2O datasets. I2S demonstrates state-of-the-art performance while maintaining a lightweight model size of under 4 MB and a fast inference time of 0.1 seconds. These characteristics make the proposed framework highly suitable for real-time, on-device authentication in security-critical, AR-based systems.}





\maketitle

\section{Introduction}\label{sec1}

Computer vision-enabled Augmented Reality (AR) systems have seen increasing integration into high-stakes environments such as aviation ~\cite{macchiarella2005augmented, safi2023augmented} and surgery~\cite{ghaednia2021augmented, Rojas-Muoz2020}, where they support professionals by delivering personalized, context-aware assistance. These domains are inherently human-centric and structured, such as cockpits\cite{lutnyk2025context} and operation theatres\cite{andersen2017augmented}, where a finite set of objects is manipulated by multiple users during mission-critical tasks. In such scenarios, Human-Object Interaction (HOI) recognition plays a pivotal role in enabling assistive intelligence, while user identification becomes vital for ensuring personalization, access control, and operational security.

Additionally, for such a system to be practically deployed, it must meet strict requirements: it needs to be lightweight, operate in real-time, and perform efficiently on edge devices with limited computational power.

\begin{figure}[htbp]
  \centering
  \includegraphics[width=\linewidth]{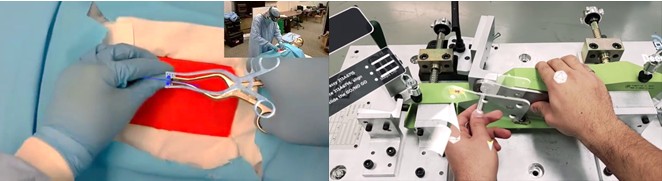}
  \caption{Augmented Reality-assisted surgery\cite{andersen2017augmented}(left) and mechanical assembly(right)}
  \label{fig:ar_headset_tasks}
\end{figure}

As illustrated in Figure~\ref{fig:ar_headset_tasks}, these AR-driven systems enhance performance and situational awareness by interpreting user actions and object interactions in real-time.

Vision-based physiological biometrics such as fingerprint \cite{Malassiotis2006}, iris \cite{daugman1994biometric}, and face recognition \cite{Deng2018} are well established. Behavioral biometrics, including gait analysis \cite{wu2016comprehensive}, touch, and keystroke dynamics \cite{Keystroke}, have also been explored for continuous authentication. However, egocentric views lack visibility of the wearer’s face or full body but prominently capture the hands, the primary medium of interaction with the environment. \cite{Tsutsui2021} were the first to explore hand gesture-based user-identification in egocentric videos.

Egocentric hand pose estimation~\cite{Oikonomidis2011,Abella2013,cai2019exploiting, fan2020adaptive, kim2021end} and HOI recognition~\cite{Tekin2019,Tsutsui2021, Liu2022, Kwon2021, Wen2022,cho2023transformer} have been extensively studied in recent years. Most of these methods leverage deep learning with diverse inputs such as 3D hand pose, 6D object pose, depth, and segmentation masks. However, these approaches are often computationally expensive, data-intensive, and lack interpretability, which presents challenges for their deployment on edge devices in AR-based assistive technologies.

To address these limitations, we propose \textit{I2S}, a multi-stage framework that incrementally enriches the feature space through object detection and HOI recognition, ultimately leading to user identification. The pipeline leverages handcrafted features in combination with machine learning models to create a lightweight and interpretable system—characteristics essential for secure, explainable biometric authentication. By incorporating HOI recognition, the framework facilitates passive, privacy-preserving user identification, making it particularly well-suited for AR co-pilot systems operating in sensitive and high-stakes environments.

\textbf{Our key contributions are as follows:} \begin{enumerate} \item A multi-stage framework for user identification through HOI recognition. \item A novel feature descriptor, \textit{Inter-Hand Spatial Envelope (IHSE)}, designed for bimanual human-object interactions in egocentric settings. \item State-of-the-art performance, achieving a 97.52\% F1-score for both HOI recognition and user identification on a custom segmented and augmented dataset derived from the ARCTIC and H2O datasets. \item Comprehensive feature analysis and ablation studies to determine the most effective descriptor combinations for this task. \end{enumerate}

The remainder of the article is organized as follows. Section II discusses related work. Section III describes the dataset, preprocessing, feature extraction and proposed multi-stage framework for user identification. Section IV summarizes the experimental evaluation whereas Section V covers the Discussion on results. Section VI presents the conclusions. 

\section{Related Work}\label{sec2}

\subsection{Biometrics}
Biometric authentication systems have traditionally relied on physiological characteristics such as fingerprints~\cite{Malassiotis2006}, iris patterns~\cite{daugman1994biometric}, facial features~\cite{Deng2018}, and palm prints. Modern techniques extend this approach through continuous authentication using behavioral patterns, offering a more unobtrusive alternative. These methods leverage signals such as keystroke dynamics~\cite{Keystroke}, mouse movements, gait~\cite{wu2016comprehensive}, and touchscreen interaction patterns~\cite{ellavarason2020touch}.

However, traditional biometric techniques are not suitable for AR co-pilots, as physiological modalities like fingerprints, iris, face, and gait are typically unobservable in egocentric views. Additionally, methods based on keystrokes and touchscreen interactions depend on continuous user engagement with a specific interface. Behavioral signatures can also vary due to fatigue, injury, or stress, reducing their reliability.

In contrast, object manipulation through hand movements remains consistently observable within egocentric perspectives, making it a promising foundation for interaction-based biometric authentication.

\subsection{Egocentric Datasets}
Recent research in the egocentric domain has led to the creation of several datasets. However, most real-world datasets, such as EPIC-Kitchens~\cite{Damen2022}, Ego4D~\cite{grauman2022ego4d}, Meccano~\cite{ragusa2021meccano}, Enigma-51~\cite{ragusa2024enigma}, provide only 2D bounding boxes, limiting their utility for fine-grained analysis.
The FPHA dataset~\cite{liu2020fpha} provides 3D hand poses along with object and action labels but is limited to single-hand interactions with rigid objects. HOI4D~\cite{Liu2022} expands on this by including scene point clouds and 3D hand joints for both rigid and only four articulated objects. The H2O dataset~\cite{Kwon2021} offers 3D hand poses and 6D object poses for two-hand interactions with rigid objects.

\begin{figure}[!htbp]
  \centering
  \includegraphics[width=0.6\linewidth]{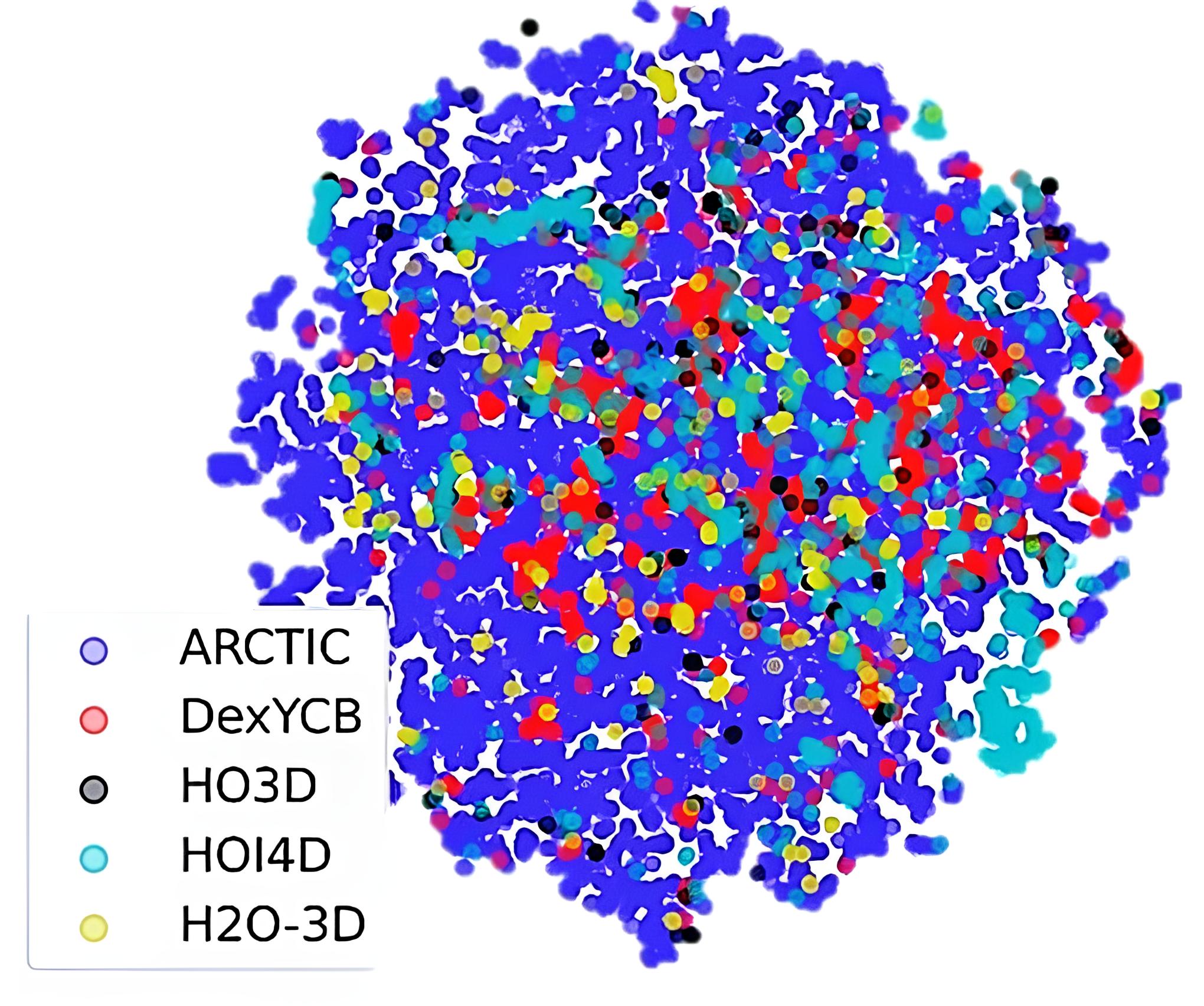}
  \caption{Comparison of hand-pose variation in egocentric datasets~\cite{Fan2022}.}
  \label{fig:handposevariation}
\end{figure}

Among the egocentric datasets, ARCTIC dataset~\cite{Fan2022} stands out by providing accurate 3D hand poses of both hands engaged in dexterous manipulation of 11 articulated objects. Figure~\ref{fig:handposevariation} illustrates the comparative complexity of hand pose variation in ARCTIC relative to other egocentric datasets.

This makes it particularly well-suited for subject-level analysis. It enables the development of a robust user identification framework relying solely on 3D hand pose estimates from egocentric videos. 

\subsection{Human-Object Interaction (HOI) Recognition in First-Person View}
HOI recognition from egocentric videos has emerged as a key research area, with multiple studies leveraging 3D hand poses to enhance activity understanding. Tekin et al.~\cite{Tekin2019} and Cho et al.~\cite{cho2023transformer} utilized 3D hand and 6D object poses with Convolutional Neural Networks and Transformer-based architectures, respectively, for HOI classification. Kwon et al.~\cite{Kwon2021} enhanced performance by incorporating depth and segmentation masks via the TA-GCN framework, while Wen et al.~\cite{Wen2022} employed a hierarchical temporal Transformer to capture the dynamics of interactions.

Despite their effectiveness in recognizing interactions, these methods do not address user identification. Moreover, their reliance on deep learning models (often requiring large datasets and significant computational resources) poses challenges for real-time deployment on AR edge devices such as headsets. Additionally, the concept of using HOI patterns as behavioral biometrics for user identification remains largely underexplored.

\subsection{Egocentric User-Identification}
Tsutsui et al.~\cite{Tsutsui2021} explored person identification using skin color, texture, and depth modalities extracted from hand gestures. Their approach showed the potential of using hands as biometric cues, but color- and texture-based modalities are highly susceptible to illumination changes and varying viewpoints, which limits reliability. Previous efforts have leveraged optical flow vectors~\cite{videoid} and gait-based cues~\cite{thapar2020sharing} to recognize individuals from egocentric videos, demonstrating that even without facial visibility, camera motion patterns and gait signatures can reliably reveal the identity of the wearer. These works highlight a critical privacy risk—namely, that egocentric video inherently encodes biometric identifiers such as head motion and gait, which can be extracted using techniques like optical flow analysis. While deep learning networks have been employed in these approaches, their dependence on large-scale training data and limited interpretability constrain their applicability in secure, on-device AR systems.

More recently, Hamid et al.\ \cite{Hamid2024} studied human-object interaction recognition using machine learning and applied in user-identification. However, their study used the HOI4D~\cite{Liu2022} dataset, which is limited to single-hand interactions and lacks sufficient subject diversity, thus limiting its applicability for implementation in AR-assisted human-centric environments.

In contrast, this study targets real-world operational environments where both hands are actively engaged in dexterous manipulation. Furthermore, object articulation is a common component of these interactions, adding to their complexity. To rigorously evaluate system robustness as the number of user classes increases, it is also essential to use datasets with greater subject diversity.

\subsection{Hand-Based Feature Extraction}

Handcrafted features provide interpretability and domain-specific insights, enabling a deeper understanding of the physical and kinematic cues that influence recognition performance. Prior work has explored frequency-domain descriptors~\cite{Abella2013}, spatial relationships between joints~\cite{Meng2016}, joint angle representations~\cite{Li2020}, and temporal motion patterns~\cite{Schell2022} to characterize hand pose and action dynamics. These approaches demonstrate that features derived from joint trajectories, spatial structure, angular configuration, and motion acceleration can capture meaningful aspects of hand-object interaction.

Building on these foundations, our study integrates spatial, frequency, angular, and temporal descriptors into a unified, interpretable framework. This hybrid representation models both the structural geometry and dynamic evolution of hand movements during interaction.

However, existing literature does not explicitly address handcrafted representations of inter-hand dynamics namely, the relative motion and coordination between both hands during object manipulation. Such dynamics may provide crucial cues about the manipulated object, the grasp type, and the nature of the human-object interaction (HOI).

\section{Methodology}\label{sec3}

Our objective is to develop a computationally efficient and edge-device-compatible technique that relies solely on 3D hand pose data extracted from egocentric videos. The proposed method performs Human-Object Interaction (HOI) recognition, followed by user identification based on the recognized interactions.

\begin{figure*}[htbp]
  \centering
  \includegraphics[width=\linewidth]{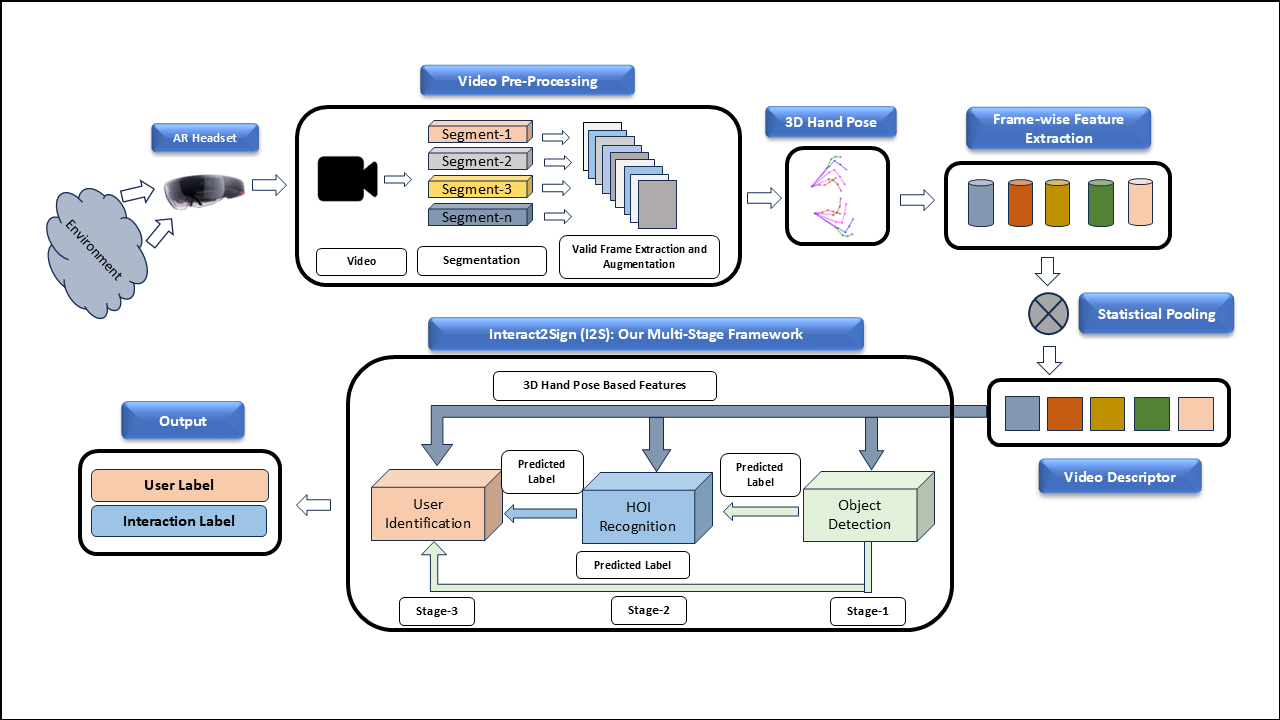}
  \caption{I2S: Multi-Stage Framework for HOIR-based User Identification}
  \label{fig:pipeline}
\end{figure*}

We introduce a multi-stage framework that sequentially executes object detection, interaction recognition, and user identification. Proposed framework is described in Figure~\ref{fig:pipeline}, and the complete algorithm is detailed in Algorithm~\ref{alg:multi_stage_pipeline}.

\begin{algorithm}
\caption{I2S: Multi-Stage HOI-based User Identification Framework}
\label{alg:multi_stage_pipeline}
\begin{algorithmic}[1]
\Function{Main}{}
  \Require 3D hand pose data extracted from egocentric videos
  \Ensure Predicted HOI and user identity $\hat{u}$

  \State \textbf{Step 1: Data Acquisition}
  \State \hspace{1em} datasets $\Leftarrow$ \Call{DataAcquisition}{}

  \State \textbf{Step 2: Segmentation and Augmentation}
  \State \hspace{1em} $\mathcal{D}_{\text{ARCTIC}} \Leftarrow$ \Call{SegmentSequences}{datasets}
  \State \hspace{1em} $\mathcal{D}_{\text{aug}} \Leftarrow \mathcal{D}_{\text{ARCTIC}} \cup \mathcal{D}_{\text{H2O}}$

  \State \textbf{Step 3: Feature Extraction}
  \ForAll{frames $f$ in $\mathcal{D}_{\text{aug}}$}
    \State $F_f \Leftarrow$ \Call{ExtractFeaturesFrom3DHandJoints}{$f$}
  \EndFor

  \State \textbf{Step 4: Feature Description}
  \ForAll{videos $v$ in $\mathcal{D}_{\text{aug}}$}
    \State $D_v \Leftarrow$ \Call{AggregateFrames}{$F_v$}
  \EndFor

  \State \textbf{Step 5: Multi-Stage Classification Pipeline}
  \State $\hat{o} \Leftarrow$ \Call{PerformObjectDetection}{$D_v$}
  \State $F^{(1)} \Leftarrow$ \Call{AugmentFeatures}{$D_v$, $\hat{o}$}
  \State $\hat{h} \Leftarrow$ \Call{PerformHOIRecognition}{$F^{(1)}$}
  \State $F^{(2)} \Leftarrow$ \Call{AugmentFeatures}{$F^{(1)}$, $\hat{h}$}
  \State $\hat{u} \Leftarrow$ \Call{PredictUser}{$F^{(2)}$}

  \State \Return $\hat{u}$
\EndFunction
\end{algorithmic}
\end{algorithm}

\subsection{Data Acquisition and Preparation}
Several egocentric datasets have been developed for studying human-object interaction (HOI) recognition, including Epic Kitchens~\cite{Damen2022}, AssemblyHands~\cite{ragusa2024enigma}, Ego4D~\cite{grauman2022ego4d}, FPHA~\cite{liu2020fpha}, and HOI4D~\cite{Liu2022}. However, based on the requirements of real-world recordings, adequate subject diversity, accurate 3D hand pose annotations, and bimanual manipulation of articulated objects for accurately modeling target AR applications, the ARCTIC dataset~\cite{Fan2022} was found to be the most suitable choice for this study.

\begin{figure*}[htbp] 
\centering 
\includegraphics[width=0.8\linewidth]{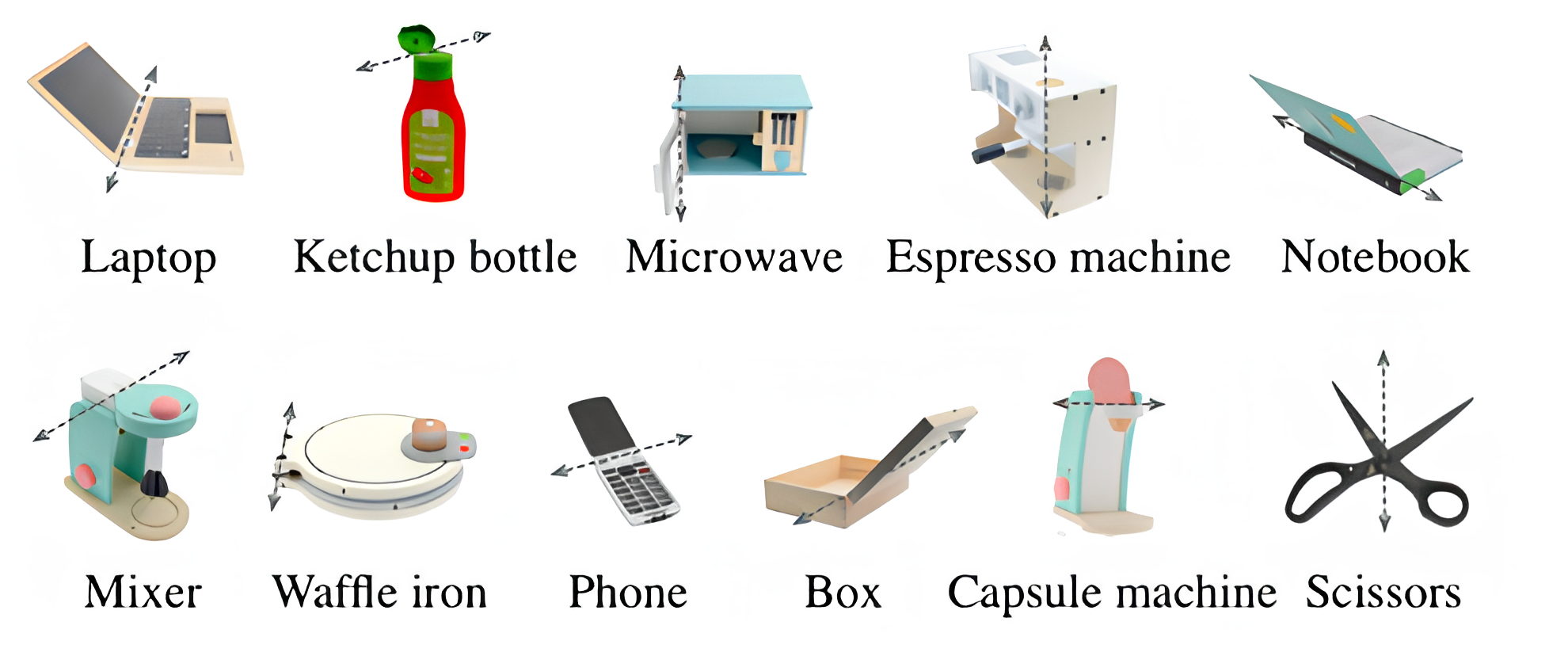} \caption{ARCTIC Objects: Each line depicts the axis of articulation~\cite{Fan2022}} \label{fig:Object_ARCTIC}
\end{figure*}

The ARCTIC dataset contains 301 videos recorded from 9 subjects interacting with 11 articulated objects depicted in Figure~\ref{fig:Object_ARCTIC}. 

Subjects in the dataset were instructed to either "grasp" or "use" the objects in a free unscripted manner. For temporal segmentation, all grasp videos were divided into 5-second segments, while use videos were segmented into 11-second windows with overlapping frames. These window durations were selected following manual observation, confirming that use interactions consistently take longer than grasp. All invalid frames were removed from videos before segmentation.

To augment the dataset, we incorporated 95 additional video segments from the H2O dataset~\cite{Kwon2021}. These segments featured bimanual interactions involving two objects, book and lotion, performed by three subjects. For consistency with ARCTIC, book interactions were relabeled as notebook (grasp/use), and lotion was relabeled as ketchup (grasp/use). The resulting augmented dataset comprises 12 subject classes, 22 HOI categories, and bimanual manipulation of 11 articulated objects.

\subsection{Feature Extraction and Description}

This study focuses on Human-Object Interaction (HOI) recognition and user identification using only 3D hand pose data obtained from egocentric video frames. Each frame provides 21 joint positions per hand in 3D space, resulting in a raw feature vector of $21 \times 3 \times 2 = 126$ values:
\begin{equation}
\label{eq:joint_structure}
J_i = \{J_i^R, J_i^L\} = \{\{j_{i1}^R, \dots, j_{i21}^R\}, \{j_{i1}^L, \dots, j_{i21}^L\}\}.
\end{equation}

\begin{figure*}[htbp]
  \centering
  \includegraphics[width=\linewidth]{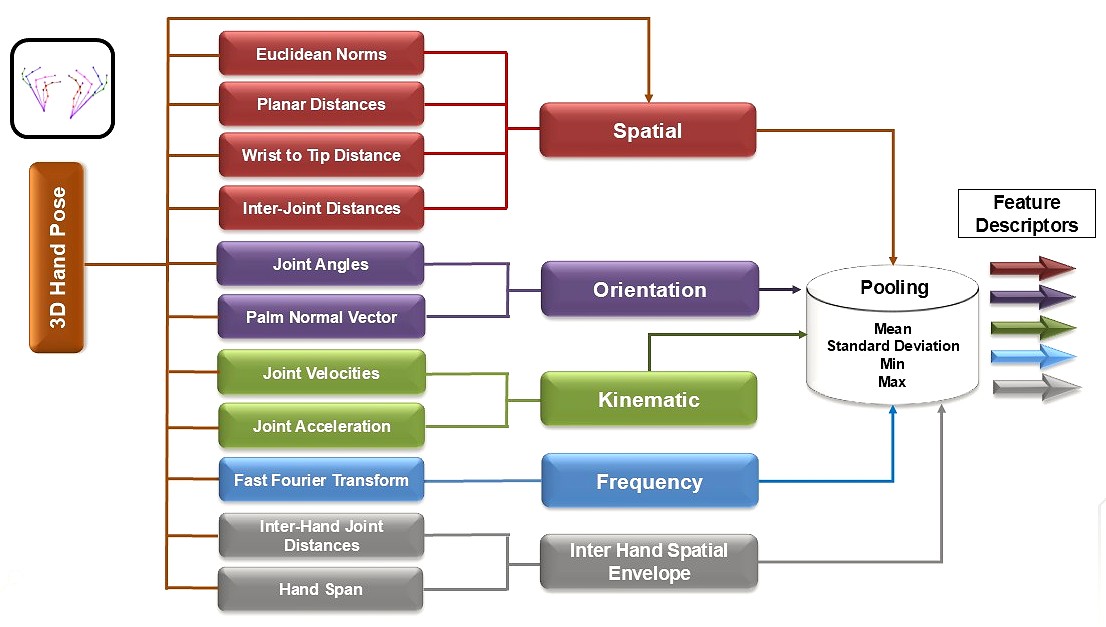}
  \caption{Feature Extraction and Description Pipeline}
  \label{fig:feature_grouping}
\end{figure*}

From this base representation, we extract and organize features into five meaningful groups—\textbf{Spatial}, \textbf{Orientation}, \textbf{Kinematic}, \textbf{Frequency-domain}, and \textbf{Inter-Hand Spatial Envelope (IHSE)}. Figure~\ref{fig:feature_grouping} describes feature extraction and description.

\subsubsection{Spatial Feature Descriptor}
Spatial features capture the intrinsic hand structure and 3D positioning. They emphasize \textit{hand posture} and \textit{shape variability} across actions. We compute:

\begin{itemize}
  \item \textbf{Euclidean Norms \cite{Hamid2024}} (42 joints x 1 = 42 features)
  \item \textbf{Planar distances \cite{Hamid2024}} (42 joints x 3 planes = 126 features)
  \item \textbf{Inter-joint distances \cite{Meng2016}} (4 inter-joints x 5 fingers x 2 hands = 40 features)
  \item \textbf{Wrist-to-tip distances} (10 finger-tips x 1 = 10 features)
  \item \textbf{3D joint coordinates} (21 joints x 3 axes x 2 hands = 126 features)
\end{itemize}

Each frame provided 344 features, which were then aggregated using Dispersion-Aware Central Tendency (DACT), i.e., mean and standard deviation, to produce the spatial video descriptor.
\begin{equation}
\label{eq:spatial_agg}
S = [\mu(x_1, \dots, x_T) \mid \sigma(x_1, \dots, x_T)] \in \mathbb{R}^{688}.
\end{equation}

\subsubsection{Orientation Feature Descriptor}
Orientation features describe the \textit{rotational posture and directionality} of the hand and palm:

\begin{itemize}
  \item \textbf{Joint angles \cite{Li2020}} (5 fingers × 3 angles × 2 hands = 30 features)
  \item \textbf{Palm normal vectors} (1 × 3 axes × 2 hands = 6 features)
\end{itemize}

These 36 features are pooled using Range-Sensitive DACT (RS-DACT), which extends DACT by including the minimum and maximum values to account for statistical extremes. This was done because hand orientation can exhibit abrupt changes across different frames of a video, making DACT alone insufficient. This enhancement ensures a more semantically accurate representation of hand orientation dynamics.

\begin{equation}
\label{eq:orient_agg}
S = [\mu, \sigma, \min, \max] \in \mathbb{R}^{144}.
\end{equation}

\subsubsection{Kinematic Feature Descriptor}

Kinematic features describe \textit{temporal dynamics\cite{Schell2022}} by modeling frame-to-frame motion:

\begin{align}
\label{eq:velocity}
v_j(t) &= \frac{p_j(t) - p_j(t-1)}{\Delta t}, \quad \Delta t = \frac{1}{30} \\
\label{eq:acceleration}
a_j(t) &= \frac{v_j(t) - v_j(t-1)}{\Delta t}, \quad \Delta t = \frac{1}{30}.
\end{align}

Given 126 3D hand joint features per frame, we compute:
\begin{itemize}
  \item \textbf{Velocities} (1 × 126 = 126 features)
  \item \textbf{Accelerations} (1 × 126 = 126 features)
\end{itemize}

These 252 features are aggregated using both central and shape-based statistics: mean ($\mu$), skewness ($\gamma_1$), and kurtosis ($\gamma_2$). Skewness and kurtosis are particularly important for distinguishing between abrupt and smooth motion patterns, which are key discriminative traits in human-object interaction dynamics.

\begin{equation}
\label{eq:kinematic_agg}
S = [\mu, \gamma_1, \gamma_2] \in \mathbb{R}^{756}.
\end{equation}

\subsubsection{Frequency-Domain Features}

To better understand the rhythm and consistency of hand motion over time, we extract features in the frequency domain using the Discrete Fourier Transform (DFT):

\begin{equation}
\label{eq:fft}
X[k] = \sum_{n=0}^{N-1} x[n] e^{-j 2 \pi nk / N}.
\end{equation}

We apply the DFT to all 126 joint coordinate sequences and compute key spectral descriptors:
\begin{itemize}
  \item \textbf{Power Spectral Density (PSD)} – reflects how motion energy is distributed across frequencies
  \item \textbf{Dominant Frequency} – identifies the most recurring motion rate
  \item \textbf{Spectral Centroid} – indicates the "center of gravity" of the frequency spectrum
  \item \textbf{Spectral Entropy} – measures complexity and unpredictability in motion
\end{itemize}

Together, these 504 features capture unique frequency patterns and periodic traits of hand-object interactions.

\subsubsection{Inter-Hand Spatial Envelope (IHSE)}

Human-object interactions in the human-centric work environments targeted by this research are predominantly bimanual, where the spatial relationship between the hands provides important cues about the object’s size, the grip used, and the type of coordination required. The IHSE feature captures this \textit{bimanual coordination}, which is essential for recognizing complex interactions:

\begin{itemize}
  \item \textbf{Hand Span} (span x 2 hands = 2 features): Measures the thumb-to-pinky distance on each hand, reflecting hand openness.
  \item \textbf{Inter-Hand Distances} (21 joints for both hands = 21 features): Computes distances between corresponding joints of the left and right hands, capturing spatial separation and synchronization.
\end{itemize}

These 23 features are aggregated using Range-Sensitive DACT (RS-DACT) to preserve central tendencies and extremes:
\begin{equation}
\label{eq:ihse_agg}
S = [\mu, \sigma, \min, \max] \in \mathbb{R}^{92}.
\end{equation}

\subsubsection{Feature Descriptor Summary}

The feature extraction pipeline yields five distinct descriptors, each tailored to capture unique physical or temporal properties of human-object interaction. These were later combined in various configurations to evaluate their individual and synergistic impacts on downstream recognition tasks.

\begin{table}[ht]
\caption{Comparison of Feature Descriptor Dimensions}
\label{tab:feature_summary}
\centering
\begin{tabular}{|l|c|}
\hline
\textbf{Feature Group} & \textbf{Descriptor Size} \\
\hline
Spatial (S) Descriptor & 688 \\
Orientation (O) Descriptor & 144 \\
Kinematic (K) Descriptor & 756 \\
Frequency-Domain (F) Descriptor & 504 \\
IHSE (I) Descriptor & 92 \\
\hline
\end{tabular}
\end{table}

\subsection{I2S: Multi-Stage Classification Framework}

This study proposes \textit{I2S}, a three-stage classification pipeline designed to progressively refine semantic understanding, starting from object recognition, advancing to interaction categorization, and culminating in subject identification. Each stage leverages the predictions from the preceding stage to enhance downstream performance. The stages are detailed as follows:

\paragraph*{\textbf{Stage 1: Object Classification}} 
An XGBoost classifier is trained on the extracted feature set to predict the object being manipulated. The predicted object label is then appended as an additional feature for the subsequent stage.

\paragraph*{\textbf{Stage 2: HOI Recognition}} 
The feature set, now augmented with the predicted object label, is input to a second XGBoost classifier trained to predict the category of the Human-Object Interaction (HOI). This step captures interaction semantics contextualized by the object class.

\paragraph*{\textbf{Stage 3: User Identification}} 
In the final stage, the feature set, further enriched with both the object and HOI predictions, is used to identify the subject performing the interaction. This stage enables personalized recognition based on the individual’s unique interaction style.

\section{Experimental Evaluation}\label{sec4}

This section presents a detailed evaluation of the proposed \textit{I2S} framework. Implementation details, experimental design, and validation strategies are reported. Our objective is to assess the effectiveness and robustness of handcrafted feature descriptors for user identification in egocentric scenarios.

\subsection{Implementation Details}

The \textit{I2S} framework employs a three-stage recognition pipeline: object classification, human-object interaction (HOI) recognition, and user identification. To facilitate this multi-stage structure, each video clip in the augmented ARCTIC dataset was annotated with object, interaction, and subject labels. These annotations serve as ground truth across all stages of classification.

Experiments were conducted using Jupyter Notebook environment on desktop with processor AMD Ryzen 5 8645HS processor, 16 GB RAM, and a 512 GB SSD. The XGBoost classifier, implemented via the \texttt{scikit-learn} library, was employed across all stages.

Multiple combinations of handcrafted feature sets, spatial, orientation, kinematic, frequency-domain, and the proposed IHSE descriptor were implemented both independently and in various fused arrangements. This allowed us to analyze how each feature group contributes to different stages of the classification pipeline. Feature fusion techniques were explored to identify optimal representations.

\subsection{Evaluation Metrics and Validation Method}
Stratified 5-Fold Cross-Validation was implemented to evaluate the performance of the framework. This ensures that class imbalances are uniformly distributed across training and test splits in every iteration, thus improving reliability and reducing evaluation bias.

Each stage, object classification, interaction recognition, and user identification, was evaluated using average F1-score metric for ablation studies. I2S's performance was calculated by averaging the performance of individual stages. 

\subsection{Performance Analysis}
We conducted an extensive ablation study across multiple descriptor groupings to evaluate their effectiveness within the I2S pipeline. Among the combinations, the \textbf{SOKI} descriptor group achieved the highest performance, yielding an overall F1-score of \textbf{97.52\%}. Table~\ref{tab:consolidatedresults} summarizes stage-wise F1-scores for all evaluated combinations, sorted in ascending order of final I2S F1-score. For brevity, descriptor group initials were concatenated (Spatial (S), Orientation (O), Kinematics (K), Frequency (F) and IHSE(I)).

\begin{table}[ht]
\centering
\caption{Ablation Study: Feature-wise F1-Scores (Sorted Top to Bottom)}
\label{tab:consolidatedresults}
\small
\begin{tabular}{lrrrr}
\toprule
\textbf{Feature Set} & \textbf{Object F1} & \textbf{HOI F1} & \textbf{Subject F1} & \textbf{I2S F1} \\
\midrule
Kinematics (K) & 55.87\% & 83.56\% & 52.83\% & 64.09\% \\
Frequency (F) & 66.02\% & 88.93\% & 74.96\% & 76.64\% \\
KF & 69.02\% & 89.92\% & 73.15\% & 77.36\% \\
KI & 83.70\% & 94.32\% & 70.23\% & 82.75\% \\
IHSE & 87.56\% & 95.28\% & 75.05\% & 85.96\% \\
KFI & 85.53\% & 95.80\% & 78.43\% & 86.58\% \\
FI & 86.50\% & 95.03\% & 78.75\% & 86.76\% \\
KFO & 91.33\% & 96.37\% & 91.25\% & 92.98\% \\
OK & 90.42\% & 96.24\% & 92.57\% & 93.08\% \\
OF & 91.47\% & 96.00\% & 93.14\% & 93.54\% \\
KOFI & 93.47\% & 97.46\% & 91.70\% & 94.21\% \\
Orientation (O) & 91.78\% & 96.80\% & 94.16\% & 94.25\% \\
SFK & 87.66\% & 95.71\% & 99.47\% & 94.28\% \\
OKI & 93.49\% & 97.47\% & 91.94\% & 94.30\% \\
SK & 87.64\% & 96.17\% & 99.64\% & 94.48\% \\
SF & 88.74\% & 95.64\% & 99.58\% & 94.65\% \\
OFI & 94.02\% & 97.02\% & 93.05\% & 94.69\% \\
Spatial (S) & 90.14\% & 95.51\% & 99.64\% & 95.10\% \\
OI & 94.43\% & 97.65\% & 93.87\% & 95.32\% \\
SOFK & 93.08\% & 96.45\% & 99.58\% & 96.37\% \\
SFI & 93.07\% & 96.94\% & 99.63\% & 96.55\% \\
SOK & 93.43\% & 97.06\% & 99.54\% & 96.68\% \\
SI & 93.74\% & 96.93\% & 99.53\% & 96.73\% \\
SFO & 93.77\% & 97.27\% & 99.58\% & 96.87\% \\
SO & 94.25\% & 97.48\% & 99.60\% & 97.11\% \\
SOFIK & 95.14\% & 97.55\% & 99.58\% & 97.42\% \\
SOI & 95.41\% & 97.51\% & 99.54\% & 97.48\% \\
SOKI & 95.16\% & 97.84\% & 99.56\% & 97.52\% \\
\bottomrule
\end{tabular}
\end{table}

\paragraph*{\textbf{Stage-1: Object Detection}}

In the object recognition stage, we observed that geometric descriptors, particularly \textit{Spatial} and \textit{Orientation}, individually offered competitive results. However, their fusion with the IHSE descriptor (denoted \textit{SOI}) yielded the best overall object detection F1-score of \textbf{95.41\%}. This notable gain underlines the complementarity of spatial arrangement and object orientation in encoding object geometry effectively.

In contrast, temporal descriptors such as velocity, acceleration, and frequency-based components showed relatively weaker discriminative power for object identity. Class-wise performance metrics 
 of \textit{SOI} descriptor for Object Classification are reported in Table~\ref{tab:object-detection-f1}. The results demonstrate consistently high accuracy across all object classes indicating strong overall model reliability. Notably, the balance between precision and recall yields robust F1-scores, suggesting the model effectively minimizes both false positives and false negatives in object recognition using SOI descriptors.

\begin{table}[htbp]
\centering
\caption{Object-wise Classification Metrics Using SOI Descriptors}
\label{tab:object-detection-f1}
\begin{tabular}{lrrrr}
\toprule
\textbf{Object Class} & \textbf{Accuracy} & \textbf{Precision} & \textbf{Recall} & \textbf{F1-score} \\
\midrule
box & 99.52\% & 94.62\% & 99.29\% & 96.84\% \\
capsule machine & 99.07\% & 95.90\% & 92.46\% & 94.07\% \\
espresso machine & 99.22\% & 96.41\% & 95.00\% & 95.64\% \\
ketchup & 98.78\% & 95.33\% & 95.20\% & 95.22\% \\
laptop & 99.08\% & 94.53\% & 95.84\% & 95.06\% \\
microwave & 99.67\% & 97.36\% & 98.79\% & 98.05\% \\
mixer & 98.95\% & 94.09\% & 94.60\% & 94.28\% \\
notebook & 98.95\% & 95.00\% & 92.38\% & 93.56\% \\
phone & 99.25\% & 95.86\% & 96.65\% & 96.22\% \\
scissors & 99.73\% & 98.70\% & 98.16\% & 98.40\% \\
waffleiron & 98.60\% & 93.03\% & 91.42\% & 92.10\% \\
\bottomrule
\end{tabular}
\end{table}

\paragraph*{\textbf{Stage-2: HOI Recognition}}
The combination of spatial, orientation, and IHSE descriptors with kinematic features (SOKI) achieved the highest performance in the HOI recognition stage, reaching an overall F1-score of 97.84\%. It is important to note that the ARCTIC dataset includes only two types of interactions, use and grasp, resulting in 22 unique object-verb pairs. This constrained interaction space enabled the models to learn highly discriminative patterns, contributing to the high accuracy. Several other feature combinations also yielded competitive results, particularly those with smaller descriptor sizes. Table~\ref{tab:hoi_f1_scores} presents the class-wise performance metrics using the SOKI feature set. 

\begin{table}[htbp]
\centering
\caption{HOI-wise performance metrics using SOKI descriptor}
\label{tab:hoi_f1_scores}
\begin{tabular}{lcccc}
\toprule
\textbf{HOI Class} & \textbf{Accuracy} & \textbf{Precision} & \textbf{Recall} & \textbf{F1-score} \\
\midrule
box\_grab & 99.79\% & 98.57\% & 95.32\% & 96.77\% \\
box\_use & 99.68\% & 93.60\% & 98.75\% & 95.96\% \\
capsulemachine\_grab & 99.90\% & 98.67\% & 98.75\% & 98.67\% \\
capsulemachine\_use & 99.84\% & 99.05\% & 97.14\% & 97.97\% \\
espressomachine\_grab & 99.76\% & 98.82\% & 96.84\% & 97.68\% \\
espressomachine\_use & 99.93\% & 100.00\% & 98.18\% & 99.05\% \\
ketchup\_grab & 99.77\% & 98.38\% & 98.18\% & 98.26\% \\
ketchup\_use & 99.78\% & 98.35\% & 98.40\% & 98.34\% \\
laptop\_grab & 99.90\% & 98.89\% & 98.89\% & 98.86\% \\
laptop\_use & 99.84\% & 97.95\% & 98.95\% & 98.41\% \\
microwave\_grab & 99.89\% & 100.00\% & 97.46\% & 98.69\% \\
microwave\_use & 99.89\% & 97.22\% & 100.00\% & 98.56\% \\
mixer\_grab & 99.68\% & 94.94\% & 99.05\% & 96.85\% \\
mixer\_use & 99.59\% & 96.43\% & 93.70\% & 94.96\% \\
notebook\_grab & 99.78\% & 95.50\% & 98.46\% & 96.86\% \\
notebook\_use & 99.65\% & 97.77\% & 94.82\% & 96.18\% \\
phone\_grab & 99.79\% & 97.27\% & 99.09\% & 98.14\% \\
phone\_use & 99.91\% & 99.05\% & 99.05\% & 99.05\% \\
scissors\_grab & 99.91\% & 100.00\% & 98.00\% & 98.95\% \\
scissors\_use & 99.95\% & 100.00\% & 98.82\% & 99.39\% \\
waffleiron\_grab & 99.83\% & 97.64\% & 98.67\% & 98.09\% \\
waffleiron\_use & 99.72\% & 97.57\% & 96.12\% & 96.79\% \\
\bottomrule
\end{tabular}
\end{table}

\paragraph*{\textbf{Stage-3: User-Identification}}
The Spatial feature descriptor demonstrated the best performance in user identification, achieving an mean F1-score of 99.64\%. As shown in Table~\ref{tab:subject_classification}, this representation consistently yielded near-perfect scores across most users, highlighting its effectiveness in capturing unique cues to identify subjects. Notably, users such as 3, 5, 7, 10, and 12 were identified with a perfect 100\% F1-score, underscoring the robustness of the spatial feature in modeling user-specific behaviors.
Although other combinations—such as the spatial-kinematic (SK) descriptor—performed comparably, and their added dimensionality offers no improvement. 

\begin{table}[htbp]
\centering
\caption{Subject-wise classification performance metrics}
\label{tab:subject_classification}
\begin{tabular}{ccccc}
\toprule
\textbf{Subject} & \textbf{Accuracy} & \textbf{Precision} & \textbf{Recall} & \textbf{F1-Score} \\
\midrule
1  & 99.89\% & 99.53\% & 99.52\% & 99.52\% \\
2  & 99.94\% & 100.00\% & 99.50\% & 99.75\% \\
3  & 100.00\% & 100.00\% & 100.00\% & 100.00\% \\
4  & 99.94\% & 99.60\% & 100.00\% & 99.80\% \\
5  & 100.00\% & 100.00\% & 100.00\% & 100.00\% \\
6  & 99.89\% & 99.55\% & 99.52\% & 99.53\% \\
7  & 100.00\% & 100.00\% & 100.00\% & 100.00\% \\
8  & 99.89\% & 99.38\% & 99.35\% & 99.35\% \\
9  & 99.94\% & 99.13\% & 100.00\% & 99.56\% \\
10 & 100.00\% & 100.00\% & 100.00\% & 100.00\% \\
11 & 99.94\% & 100.00\% & 96.67\% & 98.18\% \\
12 & 100.00\% & 100.00\% & 100.00\% & 100.00\% \\
\bottomrule
\end{tabular}
\end{table}

\subsection{Novel Feature Descriptor: IHSE}
\begin{figure*}[htbp]
  \centering
  \includegraphics[width=\linewidth]{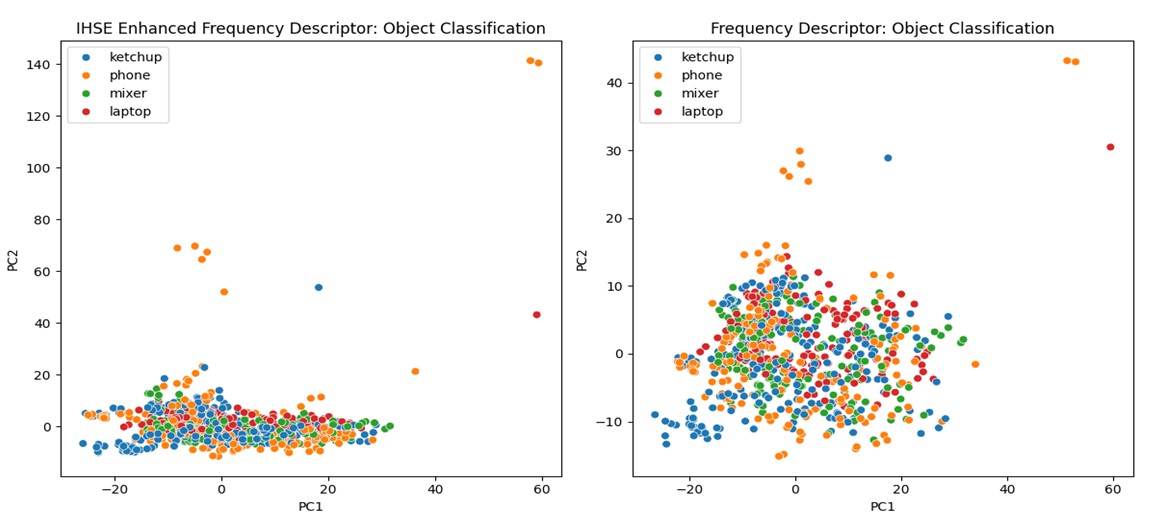}
  \caption{Comparison of PCA plots of FI vs Frequency Descriptor}
  \label{fig:FI_pca}
\end{figure*}

\begin{figure*}[htbp]
  \centering
  \includegraphics[width=\linewidth]{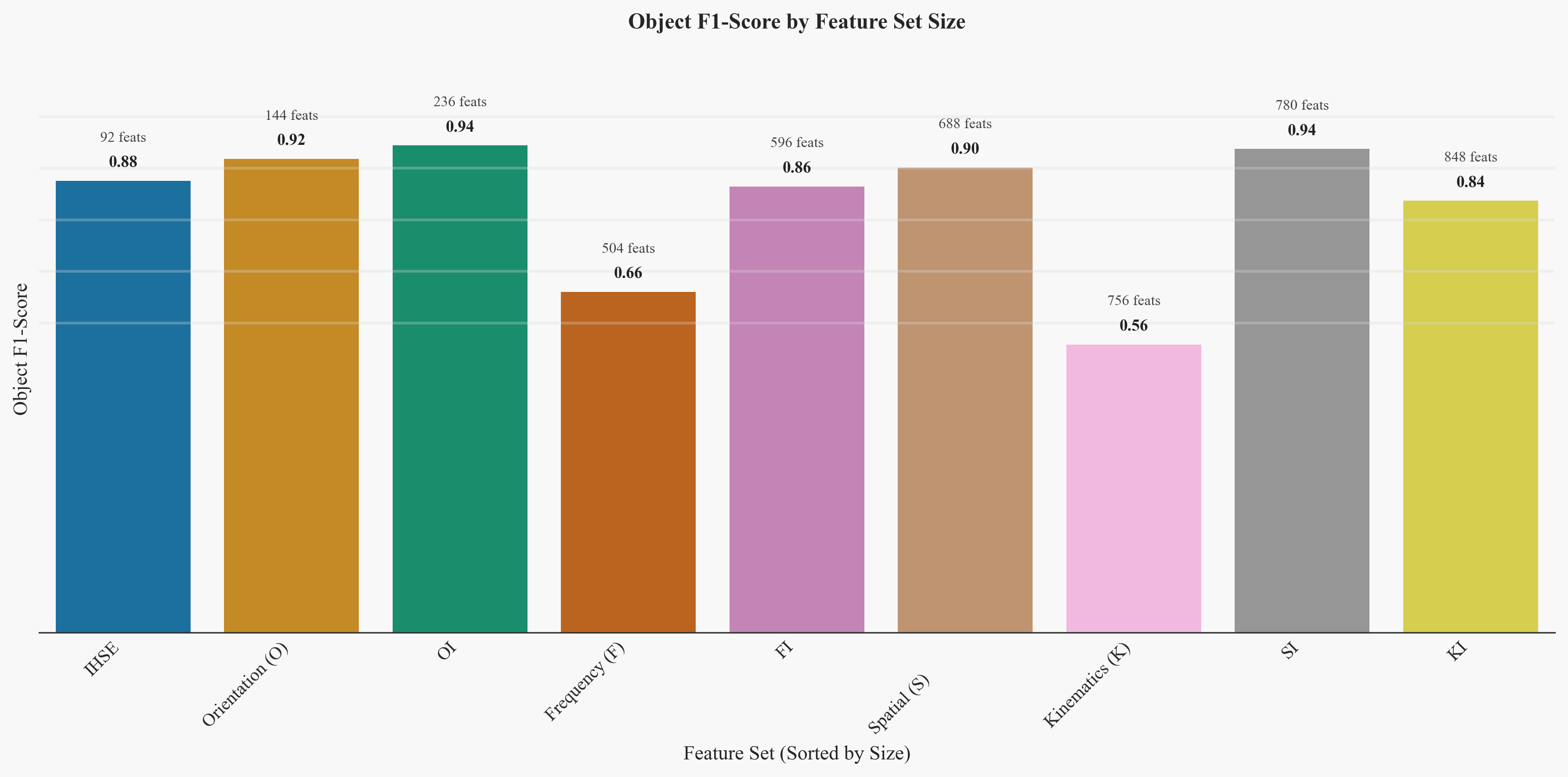}
  \caption{Enhancement of Object Detection by IHSE}
  \label{fig:IHSE_object}
\end{figure*}

\begin{figure*}[htbp]
  \centering
 \includegraphics[width=\linewidth]{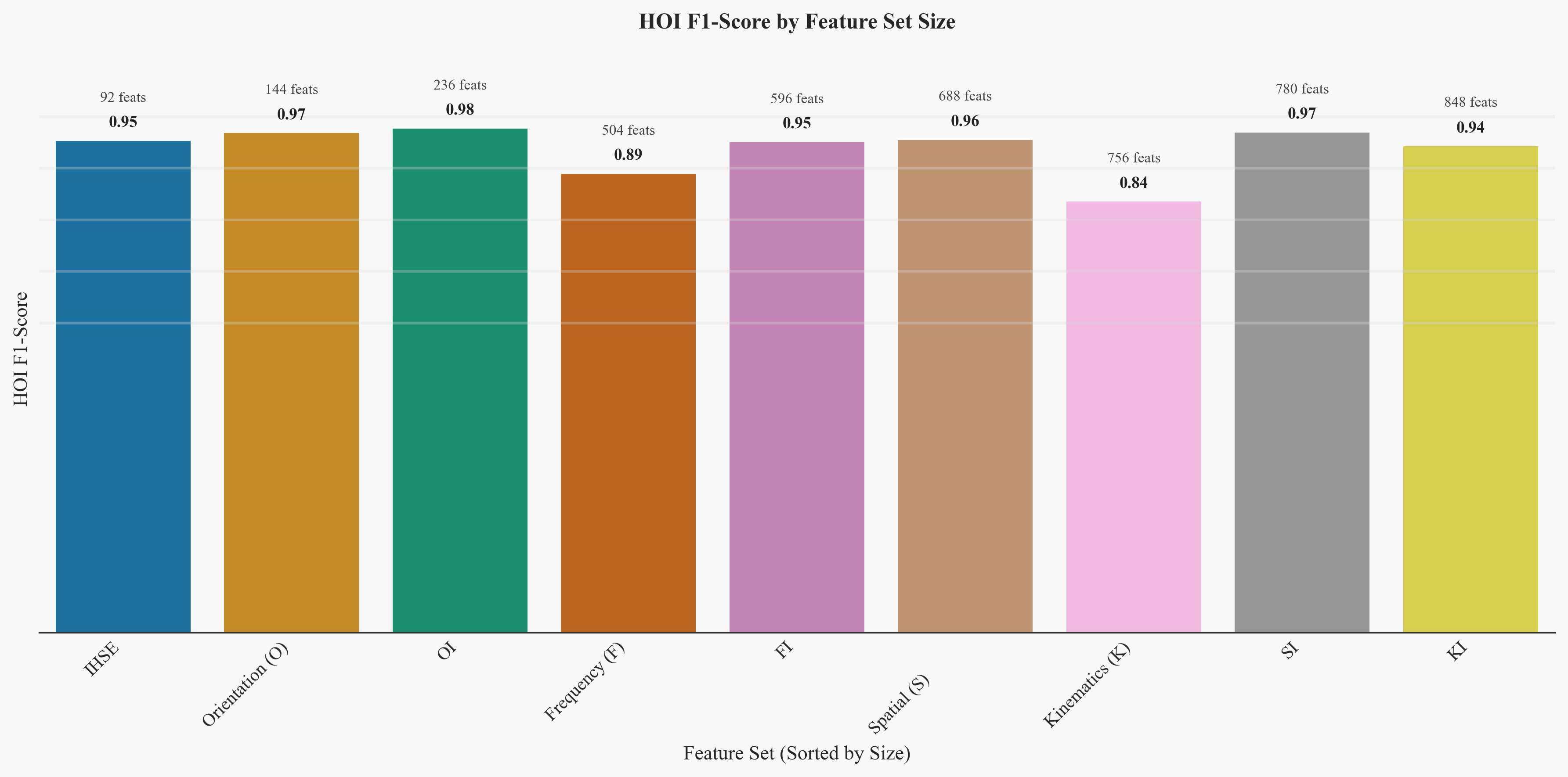}
  \caption{Enhancement of HOI Recognition by IHSE}
  \label{fig:IHSE_HOI}
\end{figure*}

Our proposed novel feature descriptor, Inter Hand Spatial Envelope, performed exceptionally well for object detection and HOI Recognition keeping in view the small feature size. Furthermore, adding IHSE to individual feature groups also enhanced their discriminative power. We perform PCA analysis on Frequency descriptor alone and in combination with IHSE and make 2D plot to compare the effect. It was found that the combined descriptor, FI, performed significantly better. Figure~\ref{fig:FI_pca} shows the PCA plots.
Figure~\ref{fig:IHSE_object} depicts the enhancement of Object Detection.
Figure~\ref{fig:IHSE_HOI} shows IHSE's impact on HOI recognition.

\subsection{Computational Analysis}
The proposed novel descriptor, IHSE, was benchmarked against the best-performing baseline descriptor, SOKI, within the I2S framework for both human-object interaction recognition and user identification. The comparative results are summarized in Table~\ref{tab:compare_soki_ihse}

\begin{table}[ht]
\centering
\caption{Comparison of SOKI and IHSE frameworks}
\label{tab:compare_soki_ihse}
\begin{tabular}{@{}lcc@{}}
\toprule
\textbf{Metric} & \textbf{SOKI} & \textbf{IHSE} \\
\midrule
\multicolumn{3}{@{}l}{\textit{Performance (F1-Score)}} \\
Object            & \textbf{95.16\%} & 87.56\% \\
HOI               & \textbf{97.84\%} & 95.28\% \\
User              & \textbf{99.56\%} & 75.05\% \\
Overall (I2S)     & \textbf{97.52\%} & 85.96\% \\
\midrule
\multicolumn{3}{@{}l}{\textit{Computational Efficiency}} \\
Training time (s) & 65.07 & \textbf{5.90} \\
Inference time (s)& 0.07  & \textbf{0.03} \\
Model size (MB)   & 3.33  & 4.05 \\
\bottomrule
\end{tabular}
\end{table}

While SOKI leads in accuracy, it demands far more time and computational resources, especially in the I2S pipeline, which takes over 65 seconds to train and 0.07 seconds to infer per sample.

IHSE, on the other hand, cuts training time by over 90\% and delivers near-instant inference (0.01–0.03s), making it highly efficient. 

Interestingly, despite using fewer features, IHSE models are slightly larger. This is due to the internal complexity of XGBoost, which builds deeper and more interconnected tree ensembles, leading to more nodes and a higher memory footprint.

Overall, IHSE trades a modest drop in F1-score for substantial gains in speed and scalability, making it ideal for real-time or edge deployments.

\subsection{Classifier Comparison}

Three different classifiers: XGBoost, Random Forest, and Support Vector Machine (SVM) were used to evaluate the performance of the I2S framework on the best feature set. Table~\ref{tab:classifier_comparison} presents a comprehensive comparison of their performance.

The results indicate that XGBoost delivers the highest overall pipeline F1-Score (97.52\%), closely followed by Random Forest (96.77\%). Interestingly, Random Forest slightly outperforms XGBoost in the object classification task with an F1-Score of 95.26\% versus 95.16\%. However, XGBoost demonstrates superior performance in HOI classification with an F1-Score of 97.84\% compared to Random Forest's 95.48\%.

Both ensemble methods (XGBoost and Random Forest) achieved identical performance in subject classification (99.56\% F1-Score), suggesting that this particular task may be relatively straightforward once accurate object and HOI classifications are established.

The SVM classifier, while still achieving respectable results, shows moderate performance degradation across all tasks. The most significant difference appears in object classification, where SVM's F1-Score (86.48\%) is approximately 9 percentage points lower than the ensemble methods. This suggests that the decision boundaries in the feature space might be more effectively captured by ensemble approaches for this particular application.

\begin{table}[ht]
\centering
\caption{Comparison of classifiers based on F1-score}
\label{tab:classifier_comparison}
\begin{tabular}{@{}lcccc@{}}
\toprule
\textbf{Classifier} & \textbf{Object} & \textbf{HOI} & \textbf{Subject} & \textbf{Framework} \\
\midrule
XGBoost       & 95.16\% & \textbf{97.84\%} & \textbf{99.56\%} & \textbf{97.52\%} \\
Random Forest & \textbf{95.26\%} & 95.48\% & \textbf{99.56\%} & 96.77\% \\
SVM           & 86.48\% & 92.43\% & 95.10\% & 91.34\% \\
\bottomrule
\end{tabular}
\end{table}

The overall pipeline F1-Score, which reflects the combined effectiveness of all classification stages, demonstrates that while SVM achieves a respectable 91.34\%, the ensemble methods provide appreciably better performance. This suggests that for complex interaction classification tasks that involve sequential predictions (object → HOI → subject), the ability of ensemble methods to capture intricate relationships in high-dimensional feature spaces offers significant advantages.

\section{Discussion}\label{sec5}

Experimental results demonstrate the effectiveness of the proposed I2S framework, with high classification F1-scores across all three stages validating the handcrafted feature descriptors and recognition pipeline.

\subsection{Effectiveness of Handcrafted Descriptors}
Spatial and Orientation features contributed most significantly to object classification, confirming that shape and positioning in first-person views are strong identity indicators. Temporal features, particularly kinematics, proved crucial for interaction detection, with high F1-scores for combinations like SOKI and SOI highlighting the synergy between static geometry and dynamic motion features.

\subsection{Feature Concatenation Benefits}
Ablation studies demonstrated that feature concatenation strategies significantly improve model generalization. The SOKI descriptor yielded the highest overall I2S F1-score (97.52\%), suggesting that the integration of spatial, orientation, kinematic, and IHSE features provides complementary information essential for accurate HOI-based subject identification. Notably, the IHSE descriptor alone also exhibited strong performance, underscoring its effectiveness as a compact, multi-domain representation.

\subsection{Stage-wise Analysis and Limitations}
Object recognition results showed spatial and orientation features effectively distinguish diverse geometries in the ARCTIC dataset. For HOI recognition, kinematics enhanced performance significantly, though the dataset's limitation to only two interaction types simplifies the classification task, highlighting the need for broader interaction vocabularies.

Subject classification benefited from combined descriptors, with spatial features encoding user-specific interaction styles, suggesting applications in personalized AI systems when traditional biometric cues are unavailable.

Despite promising results, limitations include restricted dataset size and constraints of handcrafted features versus deep learning approaches. Future work should explore hybrid models integrating handcrafted descriptors with deep representations and investigate attention-based fusion techniques to enhance cross-subject generalization.

\section{Conclusion}\label{sec6}

This study presented \textit{I2S: Interact2Sign}, a multi-stage pipeline for user identification via Human-Object Interaction analysis in egocentric videos. Our approach leverages handcrafted 3D hand pose features, spatial geometry, orientation, motion dynamics, and inter-hand spatial envelope, to perform object detection, interaction recognition, and user-identification.

Experiments on the augmented ARCTIC dataset demonstrate the efficacy of our approach, with the SOKI descriptor achieving an F1-score of 97.52\% for subject identification. Ablation studies confirm the complementary nature of our feature subsets and our concatenation strategy's robustness. Results validate that 3D hand data effectively capture interaction signatures distinguishing between users.

Limitations include simulated objects in the ARCTIC dataset and restricted interaction vocabulary. Though we employed data augmentation to address class imbalance, the constrained dataset size limits generalization.

Future work should focus on:

\begin{enumerate}
\item Expanding data collection to include diverse real-world objects, richer interaction types, and more subjects
\item Developing hybrid approaches combining handcrafted features with deep learning representations
\item Investigating real-time processing optimizations for security-focused AR/VR applications
\end{enumerate}

The promising results of \textit{I2S} demonstrate significant potential for applications in user authentication systems, personalized immersive environments, assistive technologies, and cyber security frameworks for extended reality.

\section{Conflict of Interest}\label{sec7}

On behalf of all authors, the corresponding author states that there is no conflict of interest.


\begin{thebibliography}{99}

\bibitem{macchiarella2005augmented}
N. D. Macchiarella, D. A. Vincenzi, and S. Parsons, "Augmented reality as a training medium for aviation/aerospace applications," in \textit{Proceedings of the Human Factors and Ergonomics Society Annual Meeting}, 2005.

\bibitem{safi2023augmented}
M. Safi and W. Chung, "Augmented reality uses and applications in aerospace and aviation," in \textit{Springer Handbook of Augmented Reality}, 2023.

\bibitem{ghaednia2021augmented}
H. Ghaednia, S. Han, A. Pujari, M. Lobao, and K. Singh, "Augmented and virtual reality in spine surgery: current applications and future potentials," \textit{The Spine Journal}, vol. 21, no. 11, pp. 1825–1832, 2021.

\bibitem{Rojas-Muoz2020}
E. Rojas-Muñoz, C. Lin, N. Sanchez-Tamayo, M. E. Cabrera, D. Andersen, V. Popescu, J. A. Barragan, B. Zarzaur, P. Murphy, K. Anderson, T. Douglas, C. Griffis, J. McKee, A. W. Kirkpatrick, and J. P. Wachs, "Evaluation of an augmented reality platform for austere surgical telementoring: a randomized controlled crossover study in cricothyroidotomies," \textit{npj Digital Medicine}, vol. 3, no. 1, Dec. 2020, doi: 10.1038/s41746-020-0284-9.

\bibitem{lutnyk2025context}
M. Lutnyk, T. Weber, and S. Sonntag, "Context-sensitive augmented reality assistance in the cockpit," \textit{International Journal of Human–Computer Interaction}, 2025.

\bibitem{andersen2017augmented}
D. S. Andersen, V. Popescu, S. Shanghavi, B. Mullis, and J. Wachs, "An augmented reality-based approach for surgical telementoring in austere environments," \textit{Military Medicine}, vol. 182, no. S1, pp. 310–315, 2017.

\bibitem{Malassiotis2006}
S. Malassiotis, N. Aifanti, and M. G. Strintzis, "Personal authentication using 3-D finger geometry," \textit{IEEE Transactions on Information Forensics and Security}, vol. 1, no. 1, pp. 12-21, Mar. 2006, doi: 10.1109/TIFS.2005.863508.

\bibitem{daugman1994biometric}
J. G. Daugman, "Biometric personal identification system based on iris analysis," \textit{U.S. Patent 5,291,560}, 1994.

\bibitem{Deng2018}
J. Deng, J. Guo, J. Yang, N. Xue, I. Kotsia, and S. Zafeiriou, "ArcFace: Additive Angular Margin Loss for Deep Face Recognition," \textit{IEEE Transactions on Pattern Analysis and Machine Intelligence}, vol. 43, no. 12, pp. 3454-3466, Jan. 2018, doi: 10.1109/TPAMI.2021.3087709. Available: \url{http://arxiv.org/abs/1801.07698}, \url{http://dx.doi.org/10.1109/TPAMI.2021.3087709}.

\bibitem{wu2016comprehensive}
Z. Wu, Y. Huang, L. Wang, and X. Wang, "A comprehensive study on cross-view gait-based human identification with deep CNNs," \textit{IEEE Transactions on Pattern Analysis and Machine Intelligence (TPAMI)}, vol. 39, no. 2, pp. 209–226, 2016.

\bibitem{Keystroke}
S. Banerjee and D. Woodard, "Biometric Authentication and Identification Using Keystroke Dynamics: A Survey," \textit{Journal of Pattern Recognition Research}, vol. 7, pp. 116-139, Jan. 2012, doi: 10.13176/11.427.

\bibitem{Tsutsui2021}
S. Tsutsui, Y. Fu, and D. Crandall, "Whose hand is this? person identification from egocentric hand gestures," in \textit{Proceedings - 2021 IEEE Winter Conference on Applications of Computer Vision, WACV 2021}, 2021, pp. 3398-3407, doi: 10.1109/WACV48630.2021.00344.

\bibitem{Oikonomidis2011}
I. Oikonomidis, N. Kyriazis, and A. Argyros, "Efficient model-based 3D tracking of hand articulations using Kinect," in \textit{Proc. British Machine Vision Conference}, 2011, pp. 101.1–101.11, doi: 10.5244/c.25.101.

\bibitem{Abella2013}
J. Abella, R. Alcaide, A. Sabaté, J. Mas, S. Escalera, J. Gonzàlez, and C. Antens, "Multi-modal descriptors for multi-class hand pose recognition in human computer interaction systems," in \textit{Proc. 2013 ACM Int. Conf. on Multimodal Interaction (ICMI)}, 2013, pp. 503-508, doi: 10.1145/2522848.2532596.

\bibitem{cai2019exploiting}
Y. Cai, L. Ge, J. Cai, J. Yuan, and N. Thalmann, "Exploiting spatial-temporal relationships for 3D pose estimation via graph convolutional networks," in \textit{Proceedings of the IEEE International Conference on Computer Vision (ICCV)}, 2019.

\bibitem{fan2020adaptive}
H. Fan, P. Liu, and S. Wang, "Adaptive computationally efficient network for monocular 3D hand pose estimation," in \textit{European Conference on Computer Vision (ECCV)}, 2020.

\bibitem{kim2021end}
D. U. Kim, K. I. Kim, and S. Baek, "End-to-end detection and pose estimation of two interacting hands," \textit{Proceedings of the IEEE/CVF International Conference on Computer Vision}, pp. 11189–11198, 2021.

\bibitem{Tekin2019}
B. Tekin, F. Bogo, and M. Pollefeys, ``H+O: Unified egocentric recognition of 3D hand-object poses and interactions,'' in \textit{Proc. IEEE Conf. on Computer Vision and Pattern Recognition (CVPR)}, 2019, pp.~4506--4515.\\
doi: \texttt{10.1109/CVPR.2019.00464}


\bibitem{Liu2022}
Y. Liu, Y. Liu, C. Jiang, K. Lyu, W. Wan, H. Shen, B. Liang, Z. Fu, H. Wang, and L. Yi, "HOI4D: A 4D egocentric dataset for category-level human-object interaction," \textit{arXiv preprint arXiv:2203.01577}, Mar. 2022, [Online]. Available: \url{http://arxiv.org/abs/2203.01577}.

\bibitem{Kwon2021}
T. Kwon, B. Tekin, J. Stuhmer, F. Bogo, and M. Pollefeys, "H2O: Two hands manipulating objects for first person interaction recognition," \textit{arXiv preprint arXiv:2104.11181}, Apr. 2021, [Online]. Available: \url{http://arxiv.org/abs/2104.11181}.

\bibitem{Wen2022}
Y. Wen, H. Pan, L. Yang, J. Pan, T. Komura, and W. Wang, "Hierarchical temporal transformer for 3D hand pose estimation and action recognition from egocentric RGB videos," \textit{arXiv preprint arXiv:2209.09484}, Sep. 2022, [Online]. Available: \url{http://arxiv.org/abs/2209.09484}.

\bibitem{cho2023transformer}
S. Cho, J. Kang, Y. Choi, H. Kim, and Y. Kim, "Transformer-based unified recognition of two hands manipulating objects," in \textit{Proceedings of the IEEE/CVF Conference on Computer Vision and Pattern Recognition (CVPR)}, 2023.

\bibitem{ellavarason2020touch}
E. Ellavarason, T. Perumal, V. Ponnusamy, and R. Thinaharan, "Touch-dynamics based behavioural biometrics on mobile devices—A review," \textit{ACM Computing Surveys}, vol. 53, no. 6, pp. 1–39, 2020.

\bibitem{Damen2022}
D. Damen, H. Doughty, G. M. Farinella, A. Furnari, E. Kazakos, J. Ma, D. Moltisanti, J. Munro, T. Perrett, W. Price, and M. Wray, "Rescaling egocentric vision: Collection, pipeline and challenges for EPIC-KITCHENS-100," \textit{International Journal of Computer Vision}, vol. 130, no. 1, pp. 33--55, Jan. 2022, doi: \href{https://doi.org/10.1007/s11263-021-01531-2}{10.1007/s11263-021-01531-2}.

\bibitem{grauman2022ego4d}
K. Grauman, A. Westbury, E. Byrne, Z. Chavis, A. Furnari, R. Girdhar, J. Hamburger, H. Jiang, M. Liu, X. Liu, \textit{et al.}, "Ego4D: Around the world in 3,000 hours of egocentric video," in \textit{Proceedings of the IEEE/CVF Conference on Computer Vision and Pattern Recognition}, 2022, pp. 18995--19012.

\bibitem{ragusa2021meccano}
F. Ragusa, A. Furnari, S. Livatino, and G. M. Farinella, "The MECCANO dataset: Understanding human-object interactions from egocentric videos in an industrial-like domain," in \textit{Proceedings of the IEEE/CVF Winter Conference on Applications of Computer Vision}, 2021, pp. 1569--1578.

\bibitem{ragusa2024enigma}
F. Ragusa, R. Leonardi, M. Mazzamuto, C. Bonanno, R. Scavo, A. Furnari, and G. M. Farinella, "Enigma-51: Towards a fine-grained understanding of human behavior in industrial scenarios," in \textit{Proceedings of the IEEE/CVF Winter Conference on Applications of Computer Vision}, 2024, pp. 4549--4559.

\bibitem{liu2020fpha}
L. Liu, W. Xu, C. Lu, and others, "FPHA-Afford: A domain-specific benchmark dataset for occluded object affordance estimation in human-object-robot interaction," in \textit{Proceedings of the IEEE International Conference on Image Processing (ICIP)}, 2020, pp. 1416--1420.

\bibitem{Fan2022}
Z. Fan, O. Taheri, D. Tzionas, M. Kocabas, M. Kaufmann, M. J. Black, and O. Hilliges, "ARCTIC: A Dataset for Dexterous Bimanual Hand-Object Manipulation," \textit{arXiv preprint arXiv:2204.13662}, 2022. [Online]. Available: \url{http://arxiv.org/abs/2204.13662}.

\bibitem{videoid}
Y. Hoshen and S. Peleg, "An Egocentric Look at Video Photographer Identity," in \textit{Proceedings of the IEEE Conference on Computer Vision and Pattern Recognition (CVPR)}, 2016, pp. 4284--4292, doi: \href{https://doi.org/10.1109/CVPR.2016.464}{10.1109/CVPR.2016.464}.

\bibitem{thapar2020sharing}
P. Thapar, C. Arora, and A. Nigam, "Is sharing of egocentric video giving away your biometric signature?" in \textit{European Conference on Computer Vision (ECCV)}, 2020.

\bibitem{Hamid2024}
D. Hamid, M. E. Ul Haq, A. Yasin, F. Murtaza, and M. A. Azam, "Using 3D Hand Pose Data in Recognizing Human–Object Interaction and User Identification for Extended Reality Systems," \textit{Information (Switzerland)}, vol. 15, no. 10, 2024, doi: \href{https://doi.org/10.3390/info15100629}{10.3390/info15100629}.

\bibitem{Meng2016}
M. Meng, H. Drira, M. Daoudi, and J. Boonaert, "Human Object Interaction Recognition Using Rate-Invariant Shape Analysis of Inter Joint Distances Trajectories," in \textit{IEEE Computer Society Conference on Computer Vision and Pattern Recognition Workshops (CVPRW)}, pp. 999--1004, Dec. 2016, doi: \href{https://doi.org/10.1109/CVPRW.2016.128}{10.1109/CVPRW.2016.128}.

\bibitem{Li2020}
Y.-L. Li, X. Liu, H. Lu, S. Wang, J. Liu, J. Li, and C. Lu, "Detailed 2D-3D Joint Representation for Human-Object Interaction," \textit{arXiv preprint arXiv:2004.08154}, Apr. 2020. [Online]. Available: \href{http://arxiv.org/abs/2004.08154}{http://arxiv.org/abs/2004.08154}.

\bibitem{Schell2022}
C. Schell, A. Hotho, and M. E. Latoschik, "Comparison of Data Representations and Machine Learning Architectures for User Identification on Arbitrary Motion Sequences," \textit{IEEE Conference on Artificial Intelligence and Virtual Reality (AIVR)}, Oct. 2022, pp. 1-8, doi: 10.1109/AIVR56993.2022.00010. [Online]. Available: \href{http://arxiv.org/abs/2210.00527}{http://arxiv.org/abs/2210.00527}.

\end{thebibliography}
\end{document}